\documentclass{vgtc}          

\pdfoutput=1\relax            
\usepackage{graphicx}         
\DeclareGraphicsExtensions{.pdf,.png,.jpg,.jpeg}

\graphicspath{{figures/}{pictures/}{images/}{./}} % where to search for the images

\usepackage{microtype}               
\PassOptionsToPackage{warn}{textcomp}
       
\usepackage{tabu}                    
\usepackage{array}
\usepackage{algorithmic}
\usepackage{amsmath}
\usepackage{amssymb}
\usepackage{booktabs}                
\usepackage{color}
\usepackage{cite}                    
\usepackage{epsfig}
\usepackage{eso-pic}
\usepackage{graphicx}
\usepackage[pagebackref=true,breaklinks=true,letterpaper=true,colorlinks,bookmarks=false]{hyperref}
\usepackage{lipsum}
\usepackage{makecell}
\usepackage{mathptmx}                
\usepackage{multirow}
\usepackage{overpic}
\usepackage{ragged2e}
\usepackage{times}
\usepackage{textcomp}                
\usepackage[para,online,flushleft]{threeparttable}

\newcommand{\figref}[1]{Fig.~\ref{#1}}
\newcommand{\tabref}[1]{Tab.~\ref{#1}}

\onlineid{0}
\vgtccategory{Research}
\vgtcinsertpkg

\title{Exploring Stereovision-Based 3-D Scene Reconstruction \\
for Augmented Reality}
\author{Guang-Yu Nie\\ %
        \parbox{1.4in}{\scriptsize \centering Beijing Institute of Technology, China}
\and Yun Liu\\ %
        \parbox{1.4in}{\scriptsize \centering Nankai University, China}
\and Cong Wang\\ %
        \parbox{1.4in}{\scriptsize \centering  China Electronics Standardization Institute, China}
\and Yue Liu\thanks{e-mail: liuyue@bit.edu.cn}\\ %
        \parbox{1.4in}{\scriptsize \centering Beijing Institute of Technology, China 
        \\ AICFVE of Beijing Film Academy, China}
\and Yongtian Wang\\ %
        \parbox{1.4in}{\scriptsize \centering Beijing Institute of Technology, China
        \\ AICFVE of Beijing Film Academy, China}
}

\abstract{
Three-dimensional (3-D) scene reconstruction is one of the
key techniques in Augmented Reality (AR),
which is related to the integration of image processing 
and display systems of complex information.
Stereo matching is a computer vision based approach
for 3-D scene reconstruction.
In this paper, we explore an improved stereo matching network,
\textbf{SLED-Net}, in which a Single Long Encoder-Decoder is 
proposed to replace the stacked hourglass network in PSM-Net 
for better contextual information learning.
We compare SLED-Net to state-of-the-art methods recently published,
and demonstrate its superior performance on Scene Flow and KITTI2015 test sets.
}

\CCScatlist{
  \CCScatTwelve{Computing methodologies}{Scene understanding}{}{};
  \CCScatTwelve{Computing methodologies}{Reconstruction}{}{};
  \CCScatTwelve{Computing methodologies}{Mixed/augmented reality}{}{}}

\begin{document}
\firstsection{Introduction}
\maketitle
3-D scene reconstruction is one of
the most critical techniques in AR,
which has been developed with substantial effort
and can be conducted by either 
traditional surveying or novel 3-D modeling systems.
Traditional remotely sensed reconstruction techniques 
include two major methods, 
i.e., airborne image photogrammetry 
and LiDAR (Light Detection And Ranging),
but these techniques suffer from such problems
as time consuming, high financial and equipment costs,
difficulty of cloud point processing, 
and so on.
To overcome the disadvantages of traditional methods,
stereo matching has been introduced into
3-D modeling systems in recent years.
%
%Stereo matching, also known as depth-map-based approach,
%is one of the major methods related to image-based modeling.
%
PSM-Net \cite{chang2018pyramid} is one of
the state-of-the-art stereo matching methods,
and contains stacked hourglass networks for cost volume regularization.
%
%in which the contextual information is critical to improve various models.
%
% \red{Contextual information can be obtained using deep models due to 
% the powerful representation ability of deep learning.}
%
%Stereo matching achieves a pixel-wise prediction,
%which needs the contextual information
%consisting of both global and local information.
%
%which can gradually refine the pixel-wise result
%by increasing the stack number.
%
%The stacked hourglass networks were firstly proposed in 
%\cite{newell2016stacked} for the task of human pose estimation
%to help recover the missing body parts (e.g., legs),
%and distinguish the ambiguous background.
%
However, such networks pay less attention to the local appearance
and are thus not suitable for stereo matching.
In this paper, we propose a \textit{Single Long Encoder-Decoder network} 
(SLED-Net) to replace the stacked hourglass network used in PSM-Net 
\cite{chang2018pyramid}.
The experimental results demonstrate the effectiveness of our new 
module in stereo matching.

\begin{figure}[!tb]
 \centering
 \includegraphics[width=\linewidth]{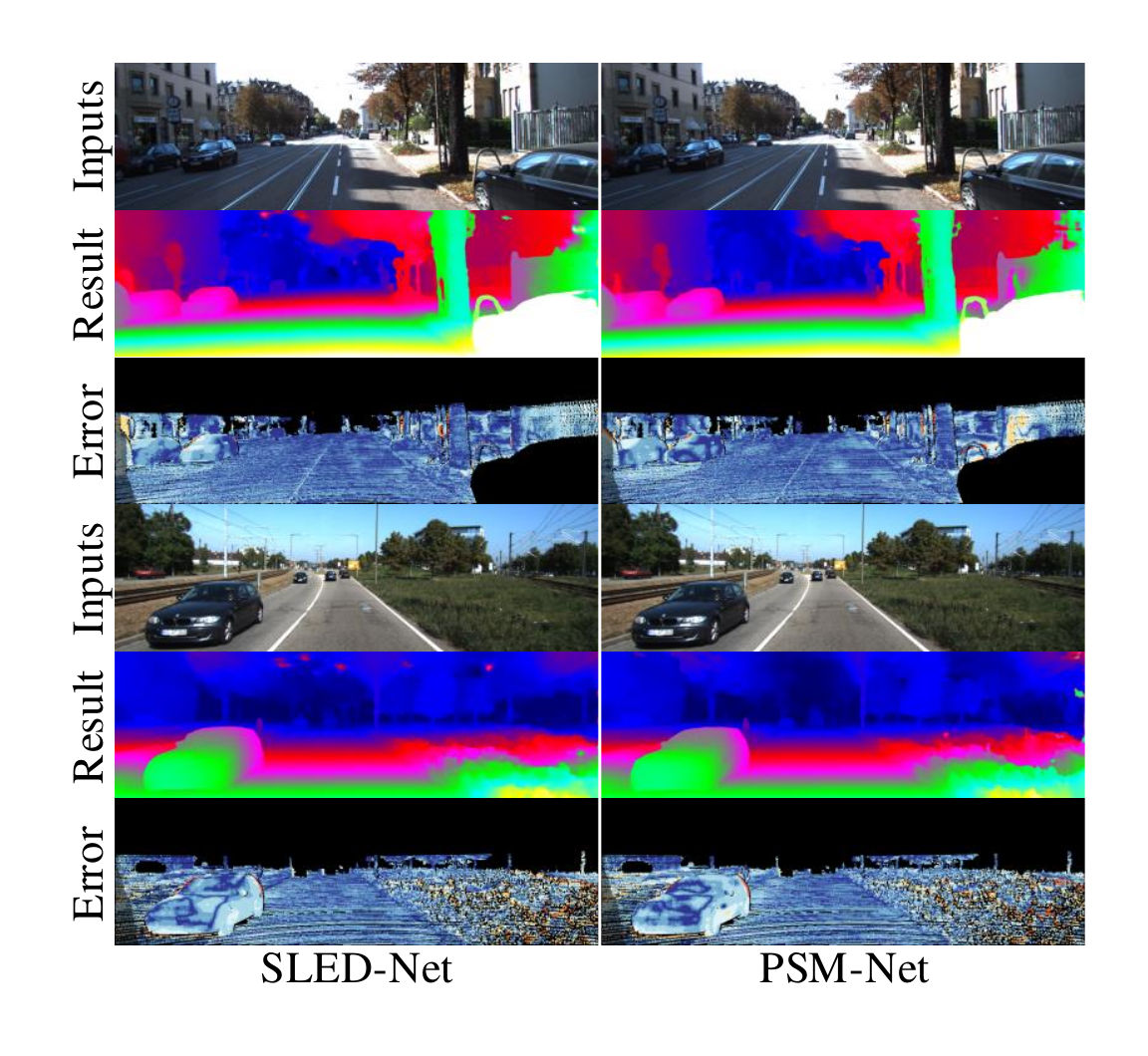}
 \vspace{-1.2cm}
 \caption{Results of our model and PSM-Net \cite{chang2018pyramid} in KITTI 2015 dataset}
 \label{fig:res2015psm}
 \vspace{-0.3cm}
\end{figure}

\begin{figure*}[!tb]
 \centering 
 \includegraphics[width=\linewidth]{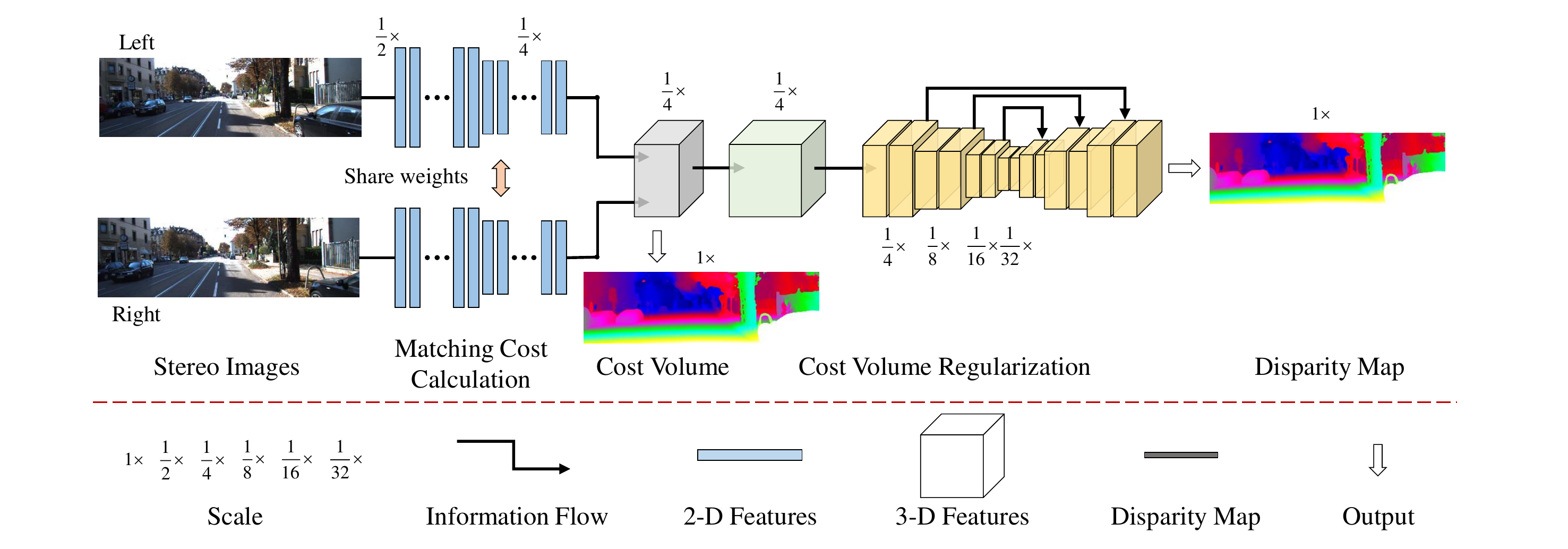}
 \vspace{-0.9cm}
 \caption{The diagrammatic sketch of SLED-Net.
          It is based on PSM-Net \cite{chang2018pyramid}, 
          but replaces the stacked hourglass networks
          %with a Single Long Encoder-Decoder 
          %(SLED, Blocks with Light Orange color).
          with SLED (Blocks with Light Orange color).
          A pair of stereo images (i.e., Left, Right) passes through
          the network to generate the disparity prediction (i.e., Disparity Map).}
 \label{fig:sample}
 \vspace{-0.4cm}
\end{figure*}

\section{Single Long Encoder-Decoder} \label{sec:SLEDarchi}
The full framework of SLED-Net is illustrated in \figref{fig:sample}.
We design a Single Long Encoder-Decoder (SLED, with light orange color)
to replace the stacked hourglass networks in PSM-Net \cite{chang2018pyramid}.
The SLED is set up as follows:
\textbf{Encoder:} Convolutional and average pooling layers
are alternatively stacked to process the feature maps down
to a resolution with $\frac{1}{32}\times$ scale
from that with $\frac{1}{4}\times$ scale.
In details, as shown in \figref{fig:sample},
the encoder totally consists of eight residual blocks,
and in each block an element-wise sum is applied to
fuse the features generated by a convolution operation
and the features generated by the previous block,
and then the fused features are feed into the subsequent block.
\textbf{Decoder:} After reaching the lowest resolution,
the network begins the top-down sequence of upsampling
and combination of features across different scales.
To bring together information across two adjacent resolutions,
we upsample the features with lower resolution 
by trilinear interpolation.
Then, we adopt an element-wise sum to combine the upsampled features
and the larger features from the encoder with skip layers,
and an operation closely following sum operation 
to fuse the combined features through a residual block 
which contains a $3\times3$ atrous convolution with dilation of 2.

To enhance the performance of SLED-Net,
the initial cost volume is used as
an intermediate supervision \cite{newell2016stacked}.
%
%The parameters of the SLED
%which is used to replace the stacked hourglass networks 
%in PSM-Net \cite{chang2018pyramid} are detailed in Tab. 1 in the Supplemental Material.

\begin{table}[!tb]
\small
\renewcommand{\arraystretch}{0.8}
\setlength\tabcolsep{3.8pt}
\caption{Performance comparison on Scene Flow Validation set}
\vspace{-0.5cm}
\begin{center}
\begin{threeparttable}
\begin{tabular}{l c | l c | l c | l c }
\toprule[1pt]
Mod. & EPE & Mod. & EPE & Mod. & EPE & Mod. & EPE \\
\midrule
\textbf{SLED-Net} &  0.699
& PSM-Net \cite{chang2018pyramid} & 1.09 
& CRL \cite{pang2017cascade} & 1.32 
& iResNet \cite{liang2018learning} & 1.40 \\
\bottomrule[1pt]
\end{tabular}
\begin{tablenotes}
\item[] \textbf{Mod.}: model;
        \textbf{EPE}: Average disparity/end-point-error.
\end{tablenotes}
\end{threeparttable}
\end{center}
\label{table:kittiscfcompare}
\vspace{-1.1cm}
\end{table}

\section{Implementation Details} \label{sec:results}
%We empirically demonstrate the performance of
%SLED-Net on several \red{standard} datasets and compare it 
%with \red{state-of-the-art competitors.}

\paragraph{Datasets}
We adopt two publicly available datasets
for training and testing:
The Scene Flow datasets % \cite{mayer2016large} 
and KITTI2015 dataset % \cite{Menze2015CVPR}.
%We evaluate the performance of our network
%on the Scene Flow validation set 
%and KITTI2015 validation and test \red{sets}. 
%\red{and then submit the results to KITTI server. (Is it not the test set???)}

\paragraph{Training}
The training of SLED-Net follows the process
described in PSM-Net \cite{chang2018pyramid}.
We implement SLED-Net using PyTorch and conduct experiments 
on four NVIDIA TITAN Xp GPUs.
%
%Before being input to the network,
%each raw image is firstly processed by color normalization
%and then randomly cropped into patches with $256 \times 512$ resolution.
%
%The network is optimized end-to-end using Adam (Adaptive Moment Estimation)
%with $\beta_1$ of 0.9 and $\beta_2$ of 0.999.
%The batch size and maximum disparity (D)
%are set to 8 and 192 pixels, respectively.
%
We first train SLED-Net on Scene Flow datasets
with a fixed learning rate of 0.001 for 20 epochs,
then we fine-tune the network on KITTI2015 dataset
with stepped learning rates of 0.001 for 600 epochs
and 0.0001 for another 400 epochs.

\section{Experimental Results}
We compare SLED-Net with three state-of-the-art approaches.
The evaluation results on the Scene Flow test set and KITTI2015 test
set are summarized in \tabref{table:kittiscfcompare} and
\tabref{table:kitti2015reschart}, respectively. 
SLED-Net achieves the superior performance on both two datasets.
Compared with PSM-Net \cite{chang2018pyramid},
although the number of parameters in SLED-Net
reduces by 0.36M,
SLED-Net surpasses PSM-Net by 35.9\% in terms of
end-point-error on Scene Flow dataset,
and by 3.9\% in terms of the overall
three-pixel-error on KITTI 2015 dataset, respectively.
We display some examples from KITTI2015 dataset in \figref{fig:res2015psm}.

\begin{table}[!tb]
\small
\renewcommand{\arraystretch}{0.8}
\setlength\tabcolsep{7.5pt}
\caption{KITTI2015 Results}
\vspace{-0.75cm}
\begin{center}
\begin{threeparttable}
\begin{tabular}{l c c c c c c}
\toprule[1pt]
\multicolumn{1}{c}{\multirow{2}*{Mod.}} & \multicolumn{3}{c}{All (\%)} & \multicolumn{3}{c}{Noc (\%)} \\ % & Runtime\\
\cmidrule(lr){2-4} \cmidrule(lr){5-7}
& D1-bg & D1-fg & D1-all & D1-bg & D1-fg & D1-all \\ % & (s) \\
\midrule
\textbf{SLED-Net}  & \textbf{1.85} & 4.15 & \textbf{2.23} & \textbf{1.70} & 3.68 & \textbf{2.02} \\
PSM-Net \cite{chang2018pyramid}   & 1.86 & 4.62 & 2.32 & 1.71 & 4.31 & 2.14 \\ % & 0.41 \\
iResNet \cite{liang2018learning} & 2.25 & \textbf{3.40} & 2.44 & 2.07 & \textbf{2.76} & 2.19 \\% & 0.12 \\
CRL \cite{pang2017cascade}   & 2.48 & 3.59 & 2.67 & 2.32 & 3.12 & 2.45 \\
\bottomrule[1pt]
\end{tabular}
\begin{tablenotes}
\item[] ``\textbf{All/Noc}'' : In total/non-occluded regions;
        ``\textbf{D1-bg/fg/all}'': Three-point error in background/foreground/all regions.
\end{tablenotes}
\end{threeparttable}
\end{center}
\label{table:kitti2015reschart}
\vspace{-1.0cm}
\end{table}

\section{Ablation Study}
To explore the reasons why SLED is better than
stacked hourglass networks in stereo matching,
we design some ablation studies:
%must demonstrate that the improvement in performance
%is not attributed to an increase in capacity
%with a larger, deeper network,
%but the change in architecture shape.
%
1) We compare PSM-Net \cite{chang2018pyramid} with a single hourglass network
(\textbf{1HG}) to a baseline network with
four Simply Cascaded Convolution layers (\textbf{SCC-Net}).
In details, the encoder of the single hourglass network in 1HG
contains four $3\times3$ convolutional layers with stride of 2,
%which processes the features down to a resolution
%with $\frac{1}{16}\times$ scale from that with $\frac{1}{4}\times$ scale.
%
SCC-Net is generated through replacing the single hourglass network in 1HG
with a plain network consisting of four $3\times3$ 
convolution layers with stride of 1.
As shown in \tabref{table:costregularization},
the performance has 26.2\% increase in Scene Flow test set 
and 6.9\% increase in KITTI2015 test set
when using a single encoder-decoder network
but with approximately the same number of layers and parameters.
2) We conduct experiments to demonstrate the stacked design
by increasing the stacked number of hourglass network from 1 to 3,
which corresponds to \textbf{1HG, 2HGs},
and \textbf{3HGs} in \tabref{table:costregularization}.
During the training the weights of all loss functions are set to 1,
which is different from such setting in PSM-Net \cite{chang2018pyramid}.
Results show that the pixel-wise estimation
can be better refined with the number of staked hourglass networks increasing.
3) We explore whether \textbf{SLED}
is better than the stacked design (\textbf{3 HGs}) in stereo matching or not.
As shown in \tabref{table:costregularization},
compared with 3HGs, the performance of SLED-Net
has 11.2\% increase in Scene Flow test set
and 1.5\% increase in KITTI2015 test set,
which demonstrates that the change in
network architecture does play the core role,
and the stacked hourglass network in PSM-Net \cite{chang2018pyramid}
limits the ability of the regularization for cost volume.

\begin{table}[!tb]
\small
\renewcommand{\arraystretch}{0.8}
\setlength\tabcolsep{6.3pt}
\caption{Ablation Study for Cost Volume Regularization}
\vspace{-0.5cm}
\begin{center}
\begin{threeparttable}
\begin{tabular}{l c c c c c c}
\toprule[1pt]
\multicolumn{1}{c}{\multirow{2}*{Mod.}} & 
\multicolumn{4}{c}{Scene Flow} & 
KITTI2015 &
\multicolumn{1}{c}{\multirow{2}*{Para.}} \\
\cmidrule(lr){2-5} \cmidrule(lr){6-6}
&  $> 1px$ & $> 3px$ & $> 5px$ & \multicolumn{1}{c}{EPE} & D1-all (\%) \\
\midrule
SCC-Net    & 12.268 & 5.213 & 3.884 & 1.273 & 2.142 & 3.84M \\
1 HG       & 9.145  & 4.045 & 2.940 & 0.939 & 1.995 & 4.06M \\
2 HGs      & 8.369  & 3.360 & 2.341 & 0.816 & 1.800 & 4.64M \\ % 496
3 HGs      & 8.335  & 3.312 & 2.303 & 0.787 & 1.754 & 5.22M \\ % 1.695
SLED-Net   & 7.044  & 3.039 & 2.098 & 0.699 & 1.728 & 4.86M \\ % 1.838
\bottomrule[1pt]
\end{tabular}
\begin{tablenotes}
\item[] $>$t\textbf{\textit{px}}: EPE larger than \textbf{t} pixels;
        \textbf{Para.}: number of parameters.
\end{tablenotes}
\end{threeparttable}
\end{center}
\label{table:costregularization}
\vspace{-0.9cm}
\end{table}

\section{Conclusion} \label{sec:conclusion}
In this paper, we present a single long encoder-decoder
to replace the stacked hourglass 
networks in PSM-Net \cite{chang2018pyramid}
when regularizing the cost volume in stereo matching.
The cost volume is encoded by several 
simply cascaded convolutional and pooling layers
and then decoded by consecutive
trilinear interpolation and $1\times1$ convolution operations
to refine the features with low resolutions.
The experimental results show that
SLED-Net achieves superior performance
on both Scene Flow datasets and KITTI2015 benchmark,
which has been demonstrated
that SLED is better than stacked design,
and SLED-Net enables to provide accurate results
for 3-D scene reconstruction in AR.

\acknowledgments{
This work has been supported by the National Key 
Research and Development Program of China (No. 2016YFB0401202) 
and the National Natural Science Foundation of China (Grant No.61731003).}

\bibliographystyle{abbrv-doi}

\bibliography{template}
\end{document}